\documentclass[runningheads]{llncs}
\usepackage[T1]{fontenc}
\usepackage{algorithm}
\usepackage{algorithmic}
\usepackage{graphicx}
\usepackage{amsmath}
\usepackage{booktabs}
\usepackage{multirow}

\usepackage{caption}
\usepackage{float}
\usepackage{placeins}
\usepackage{color, colortbl}
\usepackage{stfloats}
\usepackage{enumitem}
\usepackage{tabularx}
\usepackage{xstring}
\usepackage{xspace}
\usepackage{url}
\usepackage{subcaption}
\usepackage{xcolor}

\usepackage{verbatim}
\usepackage{amsmath}
\usepackage{amssymb}
\usepackage{bm}
\usepackage{booktabs} 
\usepackage{multirow}
\usepackage{graphicx}
\usepackage{wrapfig}
\usepackage{array}   
\usepackage{wrapfig}
\usepackage{ulem}
\usepackage{CJKulem}
\usepackage{soul}
\usepackage{hyperref}
\begin{document}
%
\title{Dwarf: Disease-weighted network for attention map refinement}

\titlerunning{Dwarf: Disease-weighted network for attention map refinement}
%
\author{Haozhe Luo\inst{1,\dagger} \and
Aurélie Pahud de Mortanges\inst{1} \and \\
Oana Inel\inst{2,*} \and Abraham Bernstein\inst{2,*} \and Mauricio Reyes\inst{1,*}}

\institute{ARTORG Center for Biomedical Engineering Research, University of Bern, Bern, Switzerland \and
University of Zurich, Switzerland}
\authorrunning{H. LUO ET AL.}
%
%
\maketitle              
\begin{abstract}

The interpretability of deep learning is crucial for evaluating the reliability of medical imaging models and reducing the risks of inaccurate patient recommendations. This study addresses the "human out of the loop" and "trustworthiness" issues in medical image analysis by integrating medical professionals into the interpretability process. We propose a disease-weighted attention map refinement network (DWARF) that leverages expert feedback to enhance model relevance and accuracy. Our method employs cyclic training \cite{ma2023foundation} to iteratively improve diagnostic performance, generating precise and interpretable feature maps. Experimental results demonstrate significant improvements in interpretability and diagnostic accuracy across multiple medical imaging datasets. This approach fosters effective collaboration between AI systems and healthcare professionals, ultimately aiming to improve patient outcomes. The code is available on \textit{censored for review}.

\end{abstract}
\section{Introduction}
Machine learning (ML) techniques, especially deep learning (DL) have significantly expanded in both research and industrial sectors, particularly with the advancements in deep neural networks (DNN). The impact and potential repercussions of these technologies have become too significant to overlook. In certain applications, failure is unacceptable; for example, a temporary malfunction in a computer vision algorithm for autonomous vehicles can result in fatalities. In the medical field, the stakes are even higher as human lives are directly affected. Early detection of diseases is crucial for patient recovery and for preventing the progression of illnesses to more severe stages. Despite recent promising results from machine learning methods \cite{ronneberger2015U,cao2023large,luo2022hybrid,you2024sarf,zhou2023learning,zhang2023knowledge,luo2024devide}, current methods are not without imperfections \cite{maier2022metrics,kaviani2022adversarial,topol2019high}. Specifically, many medical AI systems still struggle with issues such as short-cut learning \cite{geirhos2020shortcut} and misattribution \cite{hatherley2020limits}, which can hamper the reliability of medical AI system.  

The significance of interpretability in medical imaging arises from the crucial need for transparency and trust in healthcare applications of AI. Traditionally, medical imaging analysis prioritized accuracy, but with the increasing integration of AI, the emphasis has shifted towards creating understandable and explainable AI systems. The goal of explainable AI (XAI) is to make AI decision-making processes in medical imaging more comprehensible, thereby enhancing reliability and enabling healthcare professionals to effectively integrate AI tools into clinical practice \cite{van2022explainable,band2023application}. Current XAI methods interpret model outputs through various means, but due to the inherent uncertainty and complexity of deep learning patterns, translating these into intuitive interpretations for users remains difficult. This situation underscores the necessity for trustworthy ML systems in healthcare, which demand transparency and active involvement of medical professionals to ensure accuracy and relevance \cite{chen2022explainable,prentzas2023explainable}. 

Saliency map-based techniques are extensively utilized in medical explainable AI (XAI) due to their capability to highlight critical regions in medical images that influence model predictions. These techniques enhance transparency and trust in AI-driven diagnostic tools by visually representing areas of interest, such as tumors or lesions \cite{banerjee2018automated}. Methods like Grad-CAM \cite{selvaraju2017grad}, Integrated Gradients \cite{sundararajan2017axiomatic}, and SmoothGrad \cite{smilkov2017smoothgrad} provide reliable explanations by generating class-specific localization maps and reducing noise. Consequently, saliency maps play a crucial role in making AI decisions interpretable and justifiable in clinical settings, promoting their adoption in healthcare.

In this work, we tackle the "human out of the loop" and "trustworthiness" issues in medical image analysis by incorporating medical professionals into the interpretability process \cite{patricio2023explainable}. By leveraging their insights, we enhance interpretability maps, aligning deep learning explanations more closely with medical intuition. This approach enhances the relevance and utility of deep learning interpretations in medical diagnostics through expert feedback. Achieving effective human-machine teaming, where human decision-making and ML system performance are integrated, is essential for improving patient outcomes. To be specific, in this work, we are aiming to address the imperfect alignment between medical items and corresponding visual regions \cite{you2021aligntransformer}. We take the clinicians' attention annotations as visual guidance during the classification model training to simultaneously optimise attention maps and classification performance. Through extensive experiments, DWARF outperforms other baselines on both classification performance and attention map performance across different datasets. Further feedback from clinicians demonstrates that DWARF enhances clinicians' confidence in AI assisted disease classification.

\section{Method}
The object of our method is to introducing disease specific attention as a guidance during the classification model training. In this section, we introduce our DWARF from three aspects: architecture and training strategy, losses and network initialization.
\subsection{Architecture and training strategy}\label{sec:architecture}
\begin{figure}[h]
    \setlength{\belowcaptionskip}{0.1cm}
    \centering
    \includegraphics[width=13cm]{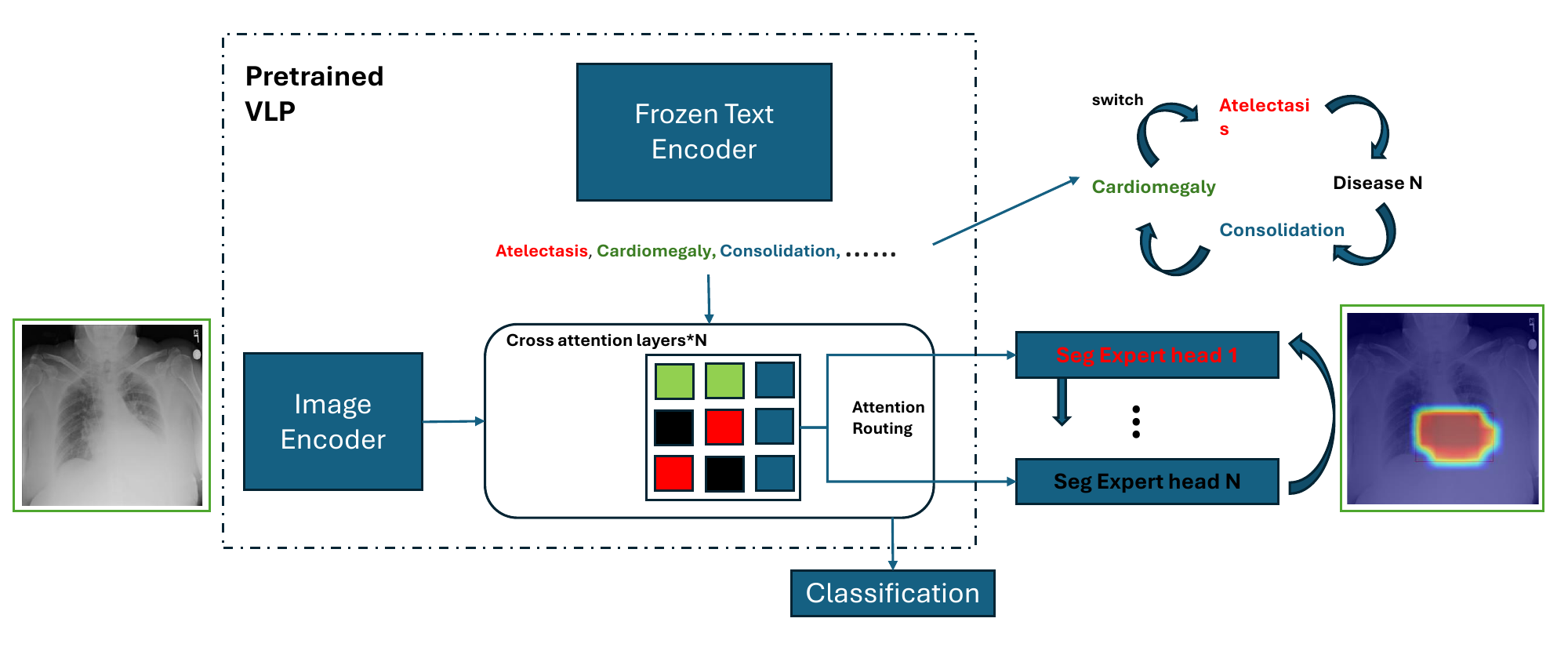}
    \caption{Flow chart of finetuning the classification model. Our method only trains single disease each epoch with disease name as prompt. For each disease, we add an additional head to mapping origin attention to refined segmentation map. }
    \label{fig:stage2} 
\end{figure}

To refine the saliency map with finding-related prior knowledge, we introduce our DWARF module, as shown in Fig.\ref{fig:stage2}. The overall structure of DWARF consists of a pretrained Vision-Language Model (VLM), denoted as \( f_{\text{vlm}} \), and expert heads \( f_{\text{heads}} \). For a multi-modality model, it is hard to directly optimize cross attention map because the attention difference between human and the model \cite{yan2023voila} as well as the scale changes (the cross attention value is not bounded within 0-1 according to experiments). To address this, we utilize the finding-specific heads to project the origin attention map denoted as \( M_c \in \mathbb{R}^{h \times w} \) of class $c$ where c is in a collection of different findings' labels $N$ from its origin embedding space to the visualization space for clinicians. Finally we get the segmentation map \( M_c' = f_{\text{head}}(M_c) \). To accumulate finding-specific knowledge effectively, we introduce a cyclic training process. The cyclic training mechanism is designed to iteratively refine the network's understanding and segmentation of specific findings. By incorporating cyclic training, the network can effectively refine its ability to identify and segment specific medical findings, leading to improved diagnostic performance. The overall training pseudo code is shown in Algorithm.\ref{Alg-Decap}

\begin{figure}[!t]

  \renewcommand{\algorithmicrequire}{\textbf{Input:}}
  \renewcommand{\algorithmicensure}{\textbf{Output:}}
  \begin{algorithm}[H]
    \caption{Training Process for DWARF}\label{Alg-Decap}
    \begin{algorithmic}[1]
      \REQUIRE Multi-label dataset $D_{multi}$, Segmentation head $f_{head}$, Ground truth $G$
      \ENSURE Optimized network parameters $\theta$
      
      \STATE \textbf{Data Collection:}
      \STATE Decompose $D_{multi}$ into multiple single-label datasets $D_{single}$
      \FOR{each finding $f$ in $D_{multi}$}
        \STATE $D_{single}[f] \leftarrow$ createSingleLabelDataset($D_{multi}$, $f$)
      \ENDFOR

      \STATE \textbf{Segmentation:}
      \FOR{each single-label dataset $D_{single}[f]$}
        \FOR{each image $I$ in $D_{single}[f]$}
          \STATE $M_c[I] \leftarrow f_{head}(I)$
        \ENDFOR
      \ENDFOR

      \STATE \textbf{Classification and Segmentation Feedback Loop:}
      \FOR{each single-label dataset $D_{single}[f]$}
        \FOR{each image $I$ in $D_{single}[f]$}
          \STATE $C_{output}[I] \leftarrow$ classify($I$, $M_c[I]$)
          \STATE $L_{seg} \leftarrow$ calculateLoss($M_c[I]$, $G[I]_{seg}$)
          \STATE $L_{cls} \leftarrow$ calculateLoss($C_{output}[I]$, $G[I]_{cls}$)
          \STATE $L_{total} \leftarrow \lambda L_{seg} + (1-\lambda)L_{cls}$
          \STATE $\theta \leftarrow$ updateNetworkParams($\theta$, $L_{total}$)
        \ENDFOR
      \ENDFOR

      \STATE \textbf{Iterative Refinement:}
      \FOR{each epoch}
        \FOR{each single-label dataset $D_{single}[f]$}
          \STATE $M_c \leftarrow$ segmentation($D_{single}[f]$, $f_{head}$)
          \STATE $\theta \leftarrow$ feedbackLoop($M_c$, $G$, $\theta$)
        \ENDFOR
      \ENDFOR
      
      \STATE Return $\theta$
    \end{algorithmic}
  \end{algorithm}
\end{figure}

\subsection{Losses and network initialization}
\subsubsection{Loss Function}
In our framework, we employ a cross-entropy loss, denoted as $\mathcal{L}_{cls}$, for multi-label classification tasks. Additionally, we use a modified Dice loss, $\mathcal{L}_{seg}$, optimized for attention maps. Attention maps in medical image analysis are critical for detecting disease-related markers. Training and validation typically focus on positive samples, which may cause models to overestimate certain features, leading to false positives. Our False Positive Suppression technique mitigates this by adjusting the Dice score to penalize false positives.

The standard metric, the Soft Dice Score, is mathematically represented as:
\begin{equation}
\mathcal{L}_{\text{Dice}} = \frac{2 \cdot |X \cap Y| + \alpha + \varepsilon}{|X| + |Y| + \alpha + \varepsilon}
\label{eq:soft_dice_score}
\end{equation}
where $X$ and $Y$ are sets representing the predicted and true regions, respectively, $\alpha$ is a smoothing constant to prevent division by zero, and $\varepsilon$ ensures numerical stability.

To specifically address false positives, we define:
\begin{equation}
\mathcal{L}_{\text{seg}} = \frac{2 \cdot |X \cap Y| + \alpha + \varepsilon}{|X| + \text{adjusted}|Y| + \alpha + \varepsilon}
\label{eq:soft_dice_score_fp}
\end{equation}
\begin{equation}
\text{adjusted}|Y| = |Y| + (w_{\text{FP}} - 1) \cdot \text{FP}
\label{eq:adjusted_cardinality}
\end{equation}
Here, $\text{FP}$ is the count of false positives, and $w_{\text{FP}}$ is the weighting factor penalizing each false positive.

The combined loss function, aimed at minimizing, is expressed as:
\begin{equation}
\mathcal{L} = \lambda \mathcal{L}_{seg} + (1-\lambda)\mathcal{L}_{cls}
\end{equation}
where $\alpha$ adjusts the emphasis on attention annotations.



\subsubsection{Model Initialization}
We initialize the text encoder using the weights from Med-KEBERT\cite{zhang2023knowledge} (An advanced text encoder pretrained on medical knowledge graph). For the image encoder and cross-attention layers, we adopt the architecture from DeViDe. Additionally, we introduce disease-specific segmentation heads for targeted diseases.

\begin{figure}[h]
    \setlength{\belowcaptionskip}{0.1cm}
    \centering
    \includegraphics[width=12cm]{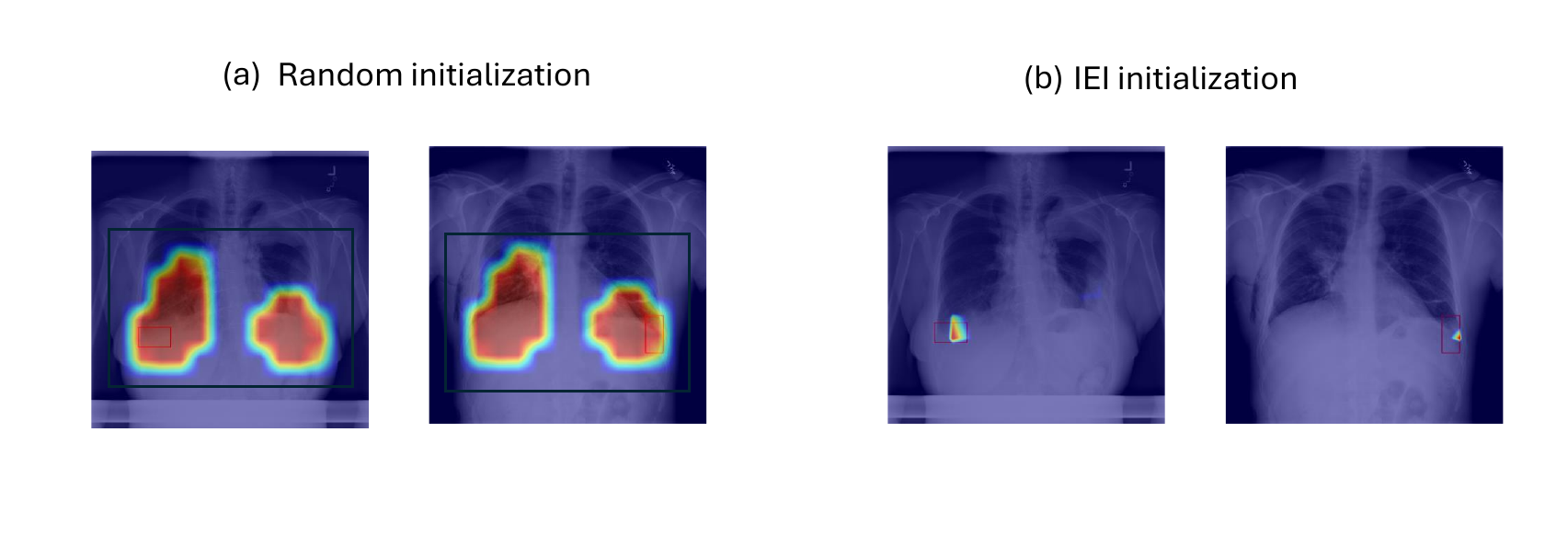}
    \caption{With random initialization, the model tends to directly learn shortcut results which always highlight the same area. While using IEI initialization, the model can start from pretrained VLM's attention to refine its focus.}
    \label{fig:iei} 
\end{figure}

We propose the \textbf{Identity Enhanced Initialization (IEI)} technique to address the limitations associated with random or simplistic initializations, which can often lead to suboptimal learning trajectories. Our observations indicate that random initialization of segmentation expert heads tends to encourage the model to learn shortcuts, as depicted in Figure~\ref{fig:iei}(a). Conversely, the pretrained Visual Language Model (VLM) already offers robust image-text correspondence\cite{luo2024devide}, which can serve as an effective foundation for initialization. The IEI method involves initializing the weights of the segmentation heads with an identity matrix, focusing on enhancing the model's sensitivity to structures pertinent to specific diseases. This approach directs the learning process towards more precise feature recognition from the start. By avoiding reliance on the simplest or most obvious features (referred to as the "shortcut path"). A qualitative comparison is illustrated in Figure~\ref{fig:iei}.

\section{Experiments and Results}
To fully assess the properties of our framework, we con- duct extensive experiments across quantitative metrics and qualitative indices. 

\subsection{Dataset}
\label{sec:dataset}
We used three different publicly available datasets: ChestX-Det \cite{liu2020chestxdet10}, CheXlocalize \cite{saporta2022benchmarking}, and Vindr-CXR \cite{nguyen2022vindr}. These datasets contain between 1,000 to 10,000 chest X-rays (CXRs). Each dataset includes multi-label classification labels as well as segmentation labels, provided at the bounding box or polygonal levels. The ChestXDet dataset is segmented into three versions based on the segmentation difficulty of the findings. The four-findings version includes common findings such as Atelectasis, Cardiomegaly, Consolidation, and Effusion, which are prevalent across various datasets. The expanded seven-findings version adds Diffuse Nodule, Emphysema, and Mass, which show relatively high performance. The full version encompasses the original ChestXDet dataset with 13 findings.

\subsection{Baselines}
To ascertain the efficacy of the DWARF method for modeling, we established several baselines for comparison. These include a pretrained vision language model without fine-tuning including DeViDe\cite{luo2024devide} and KAD\cite{zhang2023knowledge}, a finetuned VLM employing only multi-label classification loss, and a finetuned VLM training with classification loss and multi-label segmentation loss (extra supervision strategy of GAIN \cite{li2018tell}).

\subsection{Training Details }
With ViT-B as the visual backbone and Med-KEBERT as the textual backbone, we finetune on the ChestX-Det dataset \cite{lian2021structure} on an image size of 224. We utilize the AdamW optimizer with learning rates \( lr = 5 \times 10^{-5} \). We optimize on V100 16G GPUS with a total batch size of 32 for a total of 500 epochs. 

\begin{table}[hbp]
    \centering
    \setlength{\belowcaptionskip}{0.2cm}
    \caption{DWARF outperforms other finetuned/pretrained VLM on classification performance and attention accuracy on 4 metrics: AUC, Dice, F1 score and MCC. All models take the same transformer architecture as encoder. The best methods are bolded}
    \label{tab:comprehensive_comparison}
    \resizebox{1.0\columnwidth}{!}{
    \begin{tabular}{ccccccc}
    \toprule
        \textbf{Method} & \textbf{Dataset} & \textbf{AUC (\%)} & \textbf{F1 Score (\%)} & \textbf{MCC (\%)} & \textbf{Max Dice (\%)} & \textbf{Model Type} \\
    \midrule
        DeViDe \cite{luo2024devide} & ChestX-Det & 74.24 & 42.46 & 34.29 & 13.66 & Pretrained VLM \\
        KAD \cite{zhang2023knowledge} & ChestX-Det & 73.81 & 40.04 & 31.84 & 13.89 & Pretrained VLM \\
        GAIN \cite{li2018tell} & ChestX-Det & 80.90 & 48.57 & 42.65 & 13.90 & Finetuned VLM \\
        \textbf{DWARF} & ChestX-Det & $\mathbf{81.94 \pm 0.37}$ & $\mathbf{53.73 \pm 0.29}$ & $\mathbf{49.87 \pm 0.06}$ & $\mathbf{18.24 \pm 0.18}$ & Finetuned VLM  \\
    \midrule
        DeViDe & cheXlocalize & 78.26 & 41.66 & 59.83 & 11.93 & Pretrained VLM \\
        KAD & cheXlocalize & 74.22 & 58.01 & 41.53 & 11.59 & Pretrained VLM \\
        GAIN & cheXlocalize & 83.64 & 62.86 & 50.18 & 11.91 & Finetuned VLM \\
        \textbf{DWARF} & cheXlocalize & $\mathbf{84.93 \pm 0.05}$ & $\mathbf{63.44 \pm 0.21}$ & $\mathbf{50.79 \pm 0.32}$ & $\mathbf{13.40 \pm 0.51}$ & Finetuned VLM  \\
    \midrule
        DeViDe & Vindr-CXR & 72.92 & 41.28 & 31.43 & 7.19 & Pretrained VLM \\
        KAD & Vindr-CXR & 73.19 & 40.22 & 30.78 & 7.06 & Pretrained VLM \\
        GAIN & Vindr-CXR & 78.51 & 45.20 & 36.48 & 7.23 & Finetuned VLM \\
        \textbf{DWARF} & Vindr-CXR & $\mathbf{80.01 \pm 0.23}$ & $\mathbf{47.05 \pm 1.14}$ & $\mathbf{39.55 \pm 1.07}$ & $\mathbf{10.21 \pm 0.42}$ & Finetuned VLM  \\
    \bottomrule
    \end{tabular}}
\end{table}

\begin{table}[!t]
    \centering
    \setlength{\belowcaptionskip}{0.1cm}
    \caption{Comparison of DWARF and GAIN models on different numbers of selected diseases from the ChestX-Det dataset.}
    \label{tab: incremental_ablations}
    \resizebox{.95\textwidth}{!}{
    \begin{tabular}{cccccc}
    \toprule    
         Backbone & Disease number & AUC (\%) & F1 (\%) & MCC (\%) & Max Dice (\%) \\
    \midrule
        \multirow{3}*{GAIN} & 4 & 86.80 & - & - & 14.38 \\
        & 7 & 85.19 & - & - & 19.03 \\
        & 13 & 80.90 & 48.57 & 42.65 & 13.90 \\
    \midrule
         \multirow{3}*{DWARF} 
         & 4 & \textbf{88.71} & - & - & \textbf{41.47} \\
         & 7 & \textbf{87.17} & 60.17 & 52.01 & \textbf{39.11} \\
         & 13 & \textbf{81.94} & \textbf{53.73} & \textbf{49.87} & \textbf{18.24} \\
    \bottomrule
    \end{tabular}}
    
    \vspace{0.5cm}
    
    \begin{minipage}{0.48\textwidth}
    \centering
    \caption{Ablations of expert supervision. DWARF taking the trained segmentation models' generated pseudo label could also achieve impressive improvement. Seven usual findings in ChestX-Det are used for evaluation including: Atelectasis, Cardiomegaly, Consolidation, Effusion, Diffuse Nodule, Emphysema, and Mass.}
    \label{tab:expert}
    \resizebox{\textwidth}{!}{%
    \begin{tabular}{@{}lcccc@{}}
    \toprule
    Method & Disease \# & AUC (\%) & DICE (\%) & Max DICE (\%) \\ \midrule
    GaIN(cls) & 7 & 86.80 & 14.38 & 14.38 \\
    DWARF (expert) & 7 & 84.73 & 31.71 & 36.94 \\
    DWARF & 7 & \textbf{87.17} & \textbf{38.56} & \textbf{39.11} \\
    \bottomrule
    \end{tabular}
    }
    \end{minipage}%
    \hfill
    \begin{minipage}{0.42\textwidth}
    \centering
    \caption{Ablations of disease-specific head. Seven usual findings are selected for evaluation (see Table \ref{tab:expert}). Compared to directly optimizing cross attention value, introducing additional segmentation expert head improves both the classification and attention performance.}
    \label{tab:disease_specific_head}
    \resizebox{\textwidth}{!}{%
    \begin{tabular}{@{}lcc@{}}
    \toprule
    Method                    & AUC & Max Dice \\ \midrule
    Directly optimize         & 86.63   & 22.88   \\
    Disease-specific head & \textbf{87.32}   & \textbf{35.59} \\
    \bottomrule
    \end{tabular}
    }
    \end{minipage}
\end{table}

\begin{figure}[!hb]
    \centering
    \includegraphics[width=0.8\textwidth]{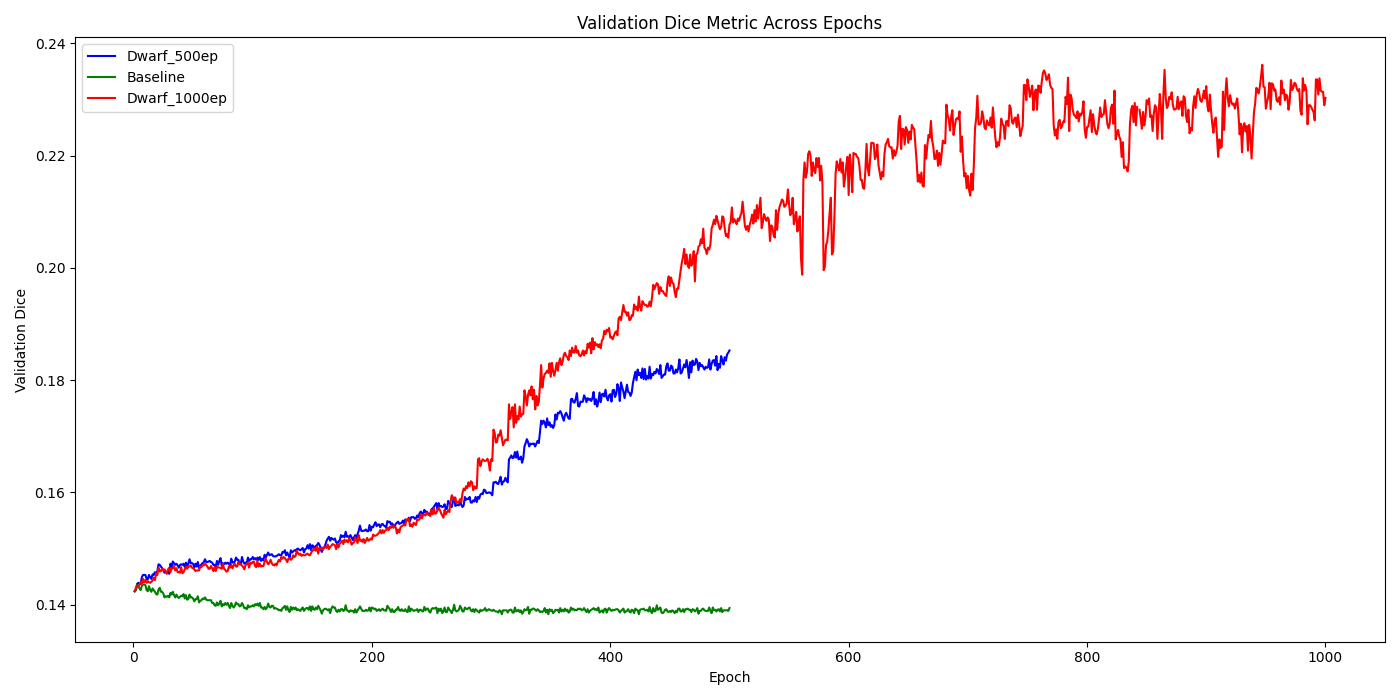}
    \caption{DWARF demonstrates sustained learning capacity, benefiting from extended training epochs, whereas the baseline model suffers from overfitting with additional training.}
    \label{fig:training_epochs}
\end{figure}

\subsection{Quantitative results}
\subsubsection{DWARF achieves SoTA results compared to other pretrained/finetuned VLM baselines.}
In the Tab \ref{tab:comprehensive_comparison}, we compared the performance of our DWARF with various state-of-the-art models, including pretrained DeViDe, KAD and finetuned GAIN which trained with direct Cross-Entropy Loss and Dice Loss. These models were evaluated based on different metrics such as Max AUC, Max Dice, F1 Score, and MCC. Our analysis extends to various datasets including ChestX-Det, cheXlocalize and Vindr-CXR, highlighting the models' adaptability and effectiveness across different medical imaging contexts. DWARF yields stably better results than the other aforementioned methods.

\subsubsection{DWARF achieves enhanced Stability and Scalability}

To explore the scalability and stability of DWARF, we firstly compared DWARF with GAIN across different numbers of diseases and observed significant and consistent improvements on ChestXDet dataset. For 4 diseases (defined in sec.\ref{sec:dataset}), the Dice score improved from 0.1438 to \textbf{0.3854}, and the max AUC score increased from 0.8660 to \textbf{0.8871}.  Similarly, for 7 diseases, the Dice score increased from 0.1903 to \textbf{0.3492}, and the max AUC score rose from 0.8519 to \textbf{0.8717}.  These results, as shown in Fig.\ref{fig:training_epochs}, highlight the consistent enhancement provided by DWARF across different numbers of diseases. Additionally, since our model is only trained once per epoch, it results in insufficient training within the same number of epochs. Therefore, we extended the training from 500 epochs to 1000 epochs to explore the scalability of the model's performance. We found that the Dice score improved from 0.1805 to 0.2302, as shown in Fig. \ref{fig:training_epochs}.

\subsubsection{DWARF's Independence from Extensive Annotation}
To address the challenges associated with obtaining dense and consistently high-quality annotations for medical imaging, we are exploring the feasibility of substituting human annotations with pseudo labels generated by disease-specific models. This approach leverages the expertise encapsulated in pre-trained segmentation models for various diseases. The results of these experiments, as detailed in Tab \ref{tab:expert}, indicate that the DWARF system continues to demonstrate substantial improvements, achieving an impressive enhancement in performance by \textbf{0.1271}.

\subsubsection{DWARF Enhances Clinician Confidence in Classification Models}
To validate our approach's effectiveness in enhancing clinicians' confidence in classification models, we conducted a double-blind experiment. We randomly selected 5 samples each of four diseases (4 common diseases across datasets: Atelectasis, Cardiomegaly, Consolidation, and Effusion) from the ChestXDet dataset, totaling 20 samples. Using DWARF and DeViDe, we generated attention maps for each sample, creating 20 anonymized sets for clinician preference evaluation. Two clinicians assessed the maps based on: 1) Accuracy (whether the map accurately pinpointed the finding with high confidence), and 2) Specificity (whether the map was focused and intuitive). DWARF was preferred in 15 out of 20 and 18 out of 20 cases, with an average preference rate of 82.5

\subsection{Ablation results}
\subsubsection{Disease-specific head makes attention map trainable}
According to the analysis presented in sec.\ref{sec:architecture}, the segmentation expert heads are pivotal for projecting the cross-attention values effectively. To explore the contributions of these heads, we conducted an ablation study, the results of which are detailed in Tab  \ref{tab:disease_specific_head}. The inclusion of expert heads significantly enhanced the attention performance, with the metric improving from 0.2288 to a robust \textbf{0.3559}.

\subsection{Qualitative Result}
\begin{figure}[h!]
    \setlength{\belowcaptionskip}{0.1cm}
    \centering
    \includegraphics[width=12cm]{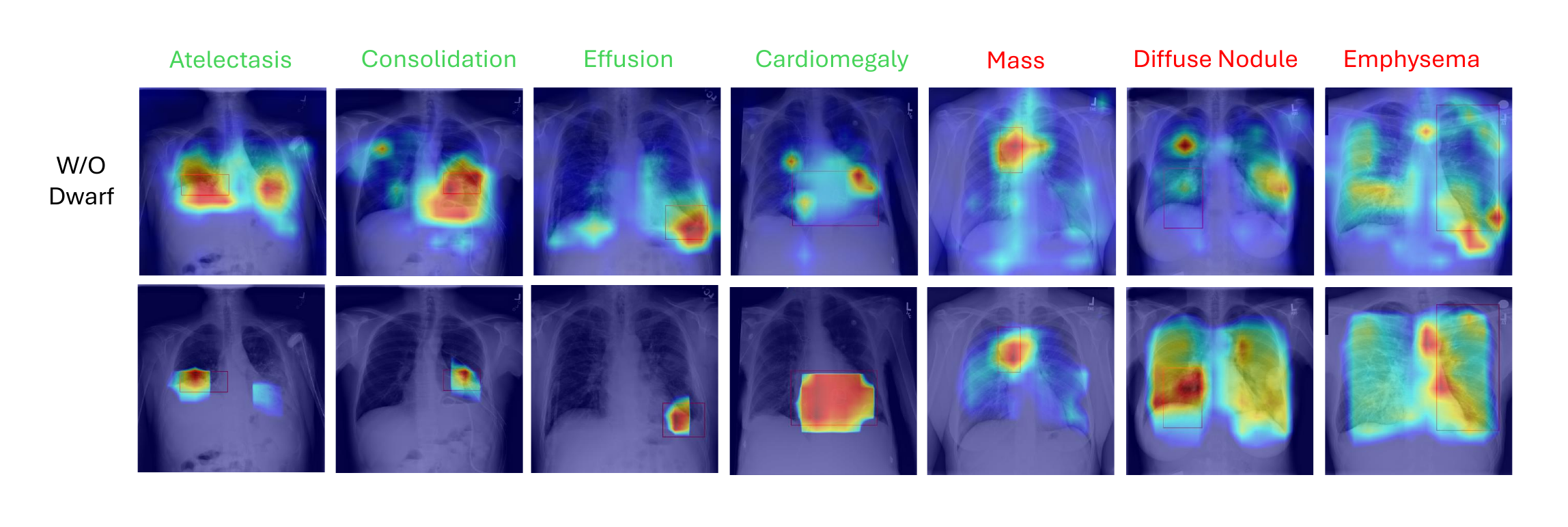}
    \caption{Qualitative results of training with and without the DWARF architecture demonstrate that utilizing our DWARF framework consistently enhances the aggregation of feature maps and provides prior region information.}
    \label{fig:qualitative_improvement} 
\end{figure}

To enhance the clarity of the explanation regarding the visual explainability representation of our method, we illustrate the attention map using the test set of the CheX-Det dataset. As depicted in Fig.\ref{fig:qualitative_improvement}, our DWARF method significantly improves the focus of the classification model's attention. This enhancement allows the model to more precisely highlight the relevant areas that form the basis for its classification decisions.

\section{Conclusion}
In this research, we have developed a two-stage saliency map revision strategy. This approach effectively integrates disease-related knowledge and clinicians' preferences into the generation of saliency maps. By incorporating this methodology, we are also introducing clinicians into the AI training loop.  This strategy not only improves the accuracy of the AI but also makes it more user-friendly for clinicians, ensuring that their expertise and insights are reflected in the AI's learning process. There are still some unvalidated capabilities  including the transferability and few-shot ability of the model. We will conduct further experiments to address them.

\FloatBarrier
\bibliographystyle{splncs04}
\bibliography{sample}
\end{document}


\title{Supplementary Material}
%
%
\author{Anonymous Submission}
\institute{Anonymous}
\authorrunning{F. Author et al.}
%
%
\maketitle              

\section{Datasets Description}

\subsection{ChestX-Det Dataset Diseases}
The ChestX-Det dataset focuses on a variety of common thoracic diseases. Below is a table listing the specific diseases included in this dataset:
\begin{table}[h]
\centering
\begin{tabular}{|c|c|}
\hline
\textbf{Disease} & \textbf{Description} \\ \hline
Atelectasis & Collapse of lung tissue affecting part or all of one lung \\ \hline
Cardiomegaly & Enlargement of the heart seen on chest X-ray \\ \hline
Effusion & Excess fluid around the lungs \\ \hline
Infiltration & The filling of airspaces with fluid and cells \\ \hline
Mass & A region in the lung seen as denser than normal \\ \hline
Nodule & A small mass of tissue in the lungs \\ \hline
Pneumonia & Infection that inflames air sacs in one or both lungs \\ \hline
Pneumothorax & Collapsed lung due to air in the chest cavity \\ \hline
Consolidation & Region of normally compressible lung tissue that has filled with liquid \\ \hline
Edema & Swelling caused by fluid in your body's tissues \\ \hline
Emphysema & A group of lung diseases that block airflow and make it difficult to breathe \\ \hline
Fibrosis & Thickening and scarring of connective tissue, usually in the lungs \\ \hline
Pleural Thickening & Thickening of the lining of the lung \\ \hline
\end{tabular}
\caption{Diseases included in the ChestX-Det Dataset}
\label{table:chestxdet}
\end{table}

\subsection{CheXlocalize Dataset Diseases}
The CheXlocalize dataset provides targeted disease localization annotations. Below is a table listing the specific diseases included in this dataset:
\begin{table}[h]
\centering
\caption{Diseases included in the CheXlocalize Dataset}
\label{table:chexlocalize}
\begin{tabular}{|l|p{0.5\textwidth}|} 
\hline
\textbf{Disease} & \textbf{Description} \\ \hline
Airspace Opacity & Areas of increased attenuation in the lungs \\ \hline
Atelectasis & Complete or partial collapse of a lung or a section (lobe) of a lung \\ \hline
Cardiomegaly & Enlargement of the heart's size, typically an indicator of heart disease \\ \hline
Consolidation & Lung tissue filled with liquid instead of air \\ \hline
Edema & Excess fluid in the lungs \\ \hline
Enlarged Cardiomediastinum & Abnormal enlargement of the mediastinal space \\ \hline
Lung Lesion & Abnormal tissue found in the lung \\ \hline
Pleural Effusion & Excess fluid that accumulates in the pleural cavity \\ \hline
Pneumothorax & Collection of air or gas in the chest or pleural space that causes part or all of a lung to collapse \\ \hline
Support Devices & Presence of devices like pacemakers, stents, etc., visible in imaging \\ \hline
\end{tabular}
\end{table}

\subsection{Vindr-CXR Dataset Diseases}
The Vindr-CXR dataset includes detailed annotations for various thoracic conditions. Below is a table listing the specific diseases included in this dataset:
\begin{table}[h]
\centering
\begin{tabular}{|l|l|}
\hline
\textbf{Disease} & \textbf{Description} \\ \hline
Aortic Enlargement & Abnormal widening or ballooning of the aorta \\ \hline
Atelectasis & Partial or complete collapse of the lung \\ \hline
Calcification & Accumulation of calcium salts in body tissue \\ \hline
Cardiomegaly & Enlargement of the heart's size due to various causes \\ \hline
Consolidation & Solidification of the lung tissue due to fluid accumulation or infection \\ \hline
ILD (Interstitial Lung Disease) & Group of diseases affecting the interstitium of the lung \\ \hline
Infiltration & Diffuse spread of inflammatory cells into lung tissues \\ \hline
Lung Opacity & Any area that appears more opaque on X-ray or scan than normal lung tissue \\ \hline
Nodule/Mass & Small abnormal but typically benign growth in the lung \\ \hline
Other lesion & Other types of abnormal lesions not classified elsewhere \\ \hline
Pleural Effusion & Excess fluid between the layers of the pleura outside the lungs \\ \hline
Pleural Thickening & Thickening of the pleural layers \\ \hline
Pneumothorax & Accumulation of air in the pleural space causing the lung to collapse \\ \hline
Pulmonary Fibrosis & Formation of excess fibrous connective tissue in the lungs \\ \hline
\end{tabular}
\caption{Diseases included in the Vindr-CXR Dataset}
\label{table:vindrcxr}
\end{table}

\section{Qualitative Results}
\begin{figure}[h!]
    \setlength{\belowcaptionskip}{0.1cm}
    \centering
    \includegraphics[width=12cm]{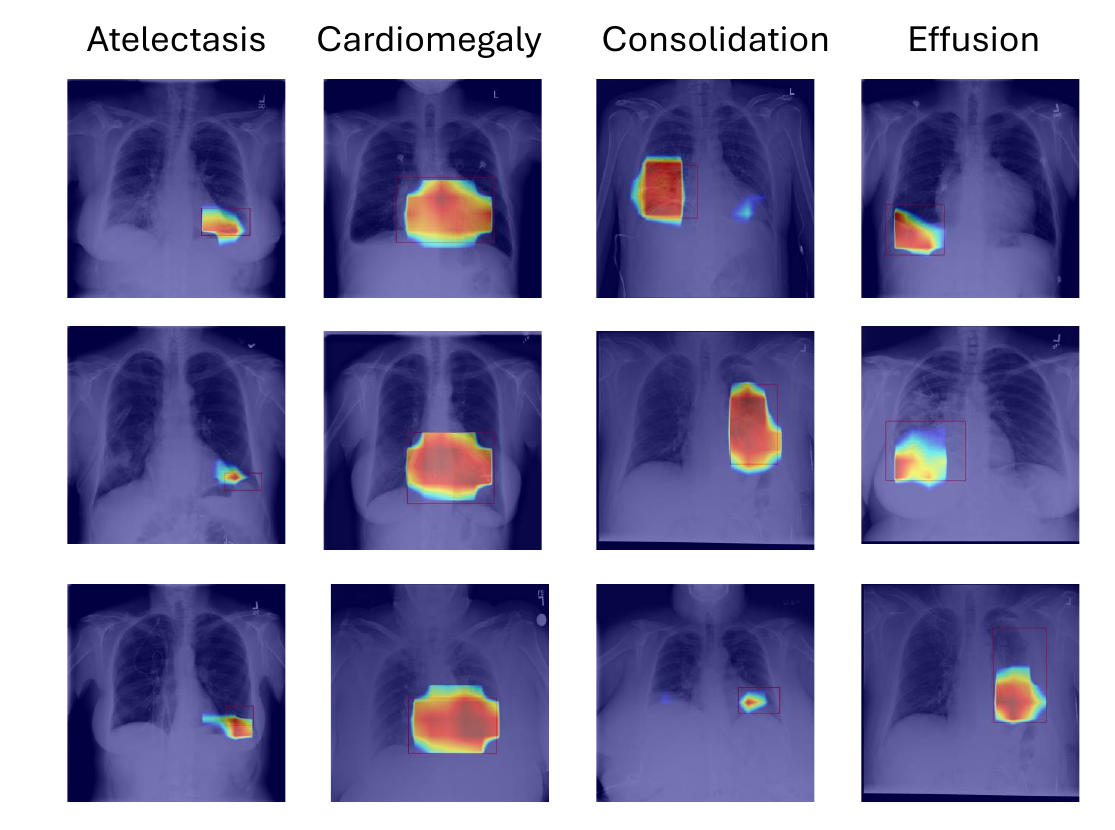}
    \caption{Qualitative results of our method. Our method could precisely locates the diseases with only attention map as explainations.}
    \label{fig:qualitative} 
\end{figure}

\begin{figure}[h!]
    \setlength{\belowcaptionskip}{0.1cm}
    \centering
    \includegraphics[width=14cm]{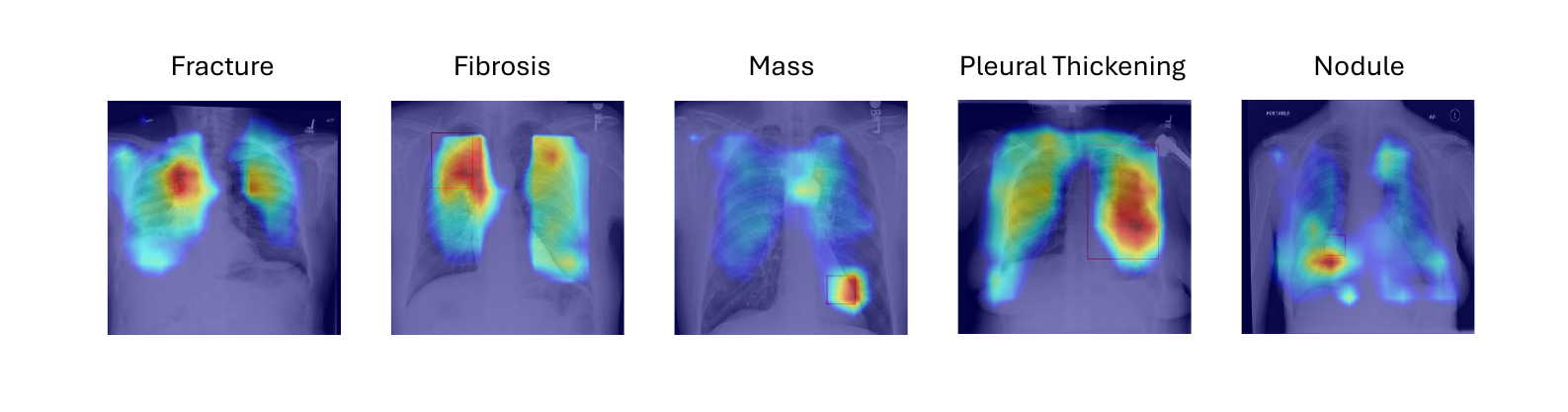}
    \caption{Qualitative results of additional findings. Even with hard to segment findings our method could still generate accurate attention maps}
    \label{fig:qualitative_further} 
\end{figure}